\documentclass[conference]{IEEEtran}
\IEEEoverridecommandlockouts
\usepackage{cite}
\usepackage{amsmath,amssymb,amsfonts}
\usepackage{algorithmic}
\usepackage{algorithm}
\usepackage{graphicx}
\usepackage{textcomp}
\usepackage{xcolor}
\DeclareUnicodeCharacter{200A}{!!!FIX ME!!!} 
\RequirePackage{graphicx,hyperref, url}

\def\BibTeX{{\rm B\kern-.05em{\sc i\kern-.025em b}\kern-.08em
    T\kern-.1667em\lower.7ex\hbox{E}\kern-.125emX}}

\DeclareRobustCommand*{\IEEEauthorrefmark}[1]{%
  \raisebox{0pt}[0pt][0pt]{\textsuperscript{\footnotesize #1}}%
}
    
\begin{document}
\title{Adaptive Batch Normalization for Training Data with Heterogeneous Features}

\author{\IEEEauthorblockN{Wael Alsobhi\IEEEauthorrefmark{1,}\IEEEauthorrefmark{4}, 
Tarik Alafif\IEEEauthorrefmark{1,}\IEEEauthorrefmark{5}, Weiwei Zong\IEEEauthorrefmark{2,}\IEEEauthorrefmark{6} and
Alaa E. Abdel-Hakim\IEEEauthorrefmark{1,}\IEEEauthorrefmark{3,}\IEEEauthorrefmark{7}}\\
\IEEEauthorrefmark{1}\textit{Computer Science Department, Umm Al-Qura University}
Jamoum, Makkah, Saudi Arabia \\
\IEEEauthorrefmark{2}\textit{WeCare.WeTeach}
Tory, Michigan, USA\\
\IEEEauthorrefmark{3}\textit{Electrical Engineering Department, Assiut University},
Assiut, Egypt, 71516\\
\\
Email: \IEEEauthorrefmark{4} s44280061@st.uqu.edu.sa,
\IEEEauthorrefmark{5} tkafif@uqu.edu.sa,
\IEEEauthorrefmark{6} wecare.weteach@gmail.com,\\
\IEEEauthorrefmark{7} adali@uqu.edu.sa, alaa.aly@eng.au.edu.eg
}

\maketitle

\begin{abstract}
Batch Normalization (BN) is an important preprocessing step to many deep learning applications. Since it is a data-dependent process, for some homogeneous datasets it is a redundant or even a performance-degrading process. In this paper, we propose an early-stage feasibility assessment method for estimating the benefits of applying BN on the given data batches. The proposed method uses a novel threshold-based approach to classify the training data batches into two sets according to their need for normalization. The need for normalization is decided based on the feature heterogeneity of the considered batch. The proposed approach is a pre-training processing, which implies no training overhead. The evaluation results show that the proposed approach achieves better performance mostly in small batch sizes than the traditional BN using MNIST, Fashion-MNIST, CIFAR-10, and CIFAR-100 datasets. Additionally, the network stability is increased by reducing the occurrence of internal variable transformation.
\end{abstract}


\section{Introduction}
Batch normalization (BN) is the process of normalizing data in a neural network to make it homogeneous. BN is introduced by Loffe and Szegedy \cite{Ioffe} to improve the classification capabilities of the
deep convolutional learning model. It standardizes the data batches with the aim of reducing the internal covariate shift. This process is typically performed by adding the batch normalization layer inside the classification model in front of the input layer. Therefore, it plays its role before entering the convolutional layers of the model.

The traditional methods apply BN blindly to all data instances in the training dataset. However, some training data do not necessarily need BN. By nature, classifiable data has small intra-class variance, and large inter-class one data. As the inter-class variance increase, the need for adaptive methods to improve classification are needed. This intuition has been applied in several data-adaptive methods, either supervised, semi-supervised~\cite{nie2017multi}, or unsupervised~\cite{hedar2018k,hedar2018modulated}. In BN, the need for normalization arises when the intra-batch distances between distances increases. BN works on reducing this large variance through mapping the data instance to a normalized feature space with respect to the original feature mean. The BN decision is usually taken in advance regardless the spread of the training data across the feature space. This may lead to a redundancy or even harm the classification accuracy. Therefore, we propose an adaptive BN approach that takes into consideration the inter-batch variations as an indicator of the normalization need prior to the training step. Small batch sizes help the effectiveness of the proposed adaptive normalization approach.

The contributions of this work can be summarized as follows:

\begin{enumerate}
	\item An adaptive BN approach is proposed to reduce the negative impacts resulted from the universal application of normalization regardless the nature of the training batches.
    \item The proposed method increases the network stability by reducing the occurrence of the internal covariate shift. 
	\item A comprehensive evaluation of the effectiveness of the proposed approach on MNIST, Fashion MNIST, CIFAR-10 and CIFAR-100 datasets is provided.
\end{enumerate}

The remainder of this paper is organized as follows: In \textbf{Section \ref{section3}},  we review related work. In \textbf{Section \ref{section2}}, preliminaries are provided. In \textbf{Section \ref{section4}}, we present the proposed adaptive BN methodology. In \textbf{Section \ref{section5}}, the conducted experiments and evaluation results are discussed. Finally, the conclusion and future work are presented in \textbf{Section \ref{section6}}.

\section{Related Work}
\label{section3}
Many research studies computed BN differently, e.g. layer normalization \cite {Jimmy}, group normalization \cite {Wu}, and instance normalization \cite{Ulyanov} based on the computing statistics on specific dimensions of the training inputs.\newline

Santurkar et al. \cite{Santurkar} presented that BN was a successful technique in optimizing the training process into a stable process. It reduced the loss and enhanced the accuracy of predictions. Also, Bjorck et al. \cite{Bjorck} concluded that the set training process parameters with optimal values, such as learning rate values, would enhance the training results. On the other hand, they claimed that the primary reason for the success of BN is its ability to enable the use of higher learning rates.

Yong et al. \cite{Yong} introduced a noise reduction BN method that was based on a momentum. In their methodology, they first revealed that the generalization capability of the BN came from its noise generation mechanism in training. Finally, they provided an improved technique built on the BN, namely momentum BN. This method used the concept of a moving average of sample mean and variance in a mini-batch for the training process. Hence, it enhances the noise level in the estimated mean and variance, which can be effectively controlled.

Sen Yang et al. \cite{Sen} proposed an adaptive self-normalization (SN). The method performed well in instance segmentation algorithm. Samples were independent of batch size to replace single BN with adaptive weight loss layer in the SN models.

Working with batches is the core concept of the BN. Li et al. \cite {Yanghao} adopted handling the normalization on a multi-batch basis. They applied data normalization over all the training samples using a pre-trained model for domain adaptation. However, their technique affects the internal covariate shift since it increases the occurrence of unstable data standardization using multiple batches for each training step.

Zhuliang et al. \cite{Zhuliang} presented a method for processing statistics with small batches, which was the cross-group iterations. The statistics were computed on a similar mass of recent iterations. To avoid the change in the weight values between similar examples, they substituted the weights based on a Taylor polynomial method. They proved the results that the method of statistical computation achieved better results than the normal BN.

Yanghao et al. \cite{Yanghao} proposed a new method for domain adaptation called adaptive BN. They added another task in the batch layer, which added the neuron value with the normalization computation without adding a parameter. The layer standardization ensured that each layer received data from the same domain. The method was proved to be effective in cloud detection of remote sensing images.

Huang et al. \cite{Huang} introduced a decorrelated BN, which extended the working of the BN by decorrelating data feature vectors using the covariance matrix. The covariance matrix contained the data standardization information and had been computed over a mini-batch differently from the normal BN. It simply centered and scaled activations. The decorrelated BN improved the performance of the BN on multilayer perceptrons and CNN since it didn't only rely on the covariance matrix to lead the normalization process but it also used the data correlation method to identify the standardization values between the data features.

Gao et al. \cite{Gao} suggested adding a simple scheme for attribute calibration by attribute positioning. It showed to be effective for the BN layer with the goal of expanding datasets and operations. They intended to enhance instance-specific representations at a minimum computational cost. They set a minimum computational cost because reliance on strong features reduces annoying features.

Chai et al. \cite{Chai} used key concepts from the field of traditional adaptive filtering by using the BN in-target to gain insight into the dynamics and inner workings of BN. They improved the behavior of the BN equations to help stabilize the BN work.

Soham et al. \cite{Soham} provided a blueprint that trained the rest of the deep network without normalization. Finally, authors in \cite{Sergey Ioffe} \cite{Benz} \cite {Huang L} rectified statistics and statistical parameters to overcome different batches.
 
All of these BN methods apply normalization blindly without taking into account the actual need of this data in this layer. Since, this blind application of BN may negatively affect the classification performance, we propose a novel adaptive BN approach.

\section{Preliminaries}
\label{section2} Loffe and Szegedy \cite{Ioffe} suggested normalizing the input data of all constructed sub-networks. Since the normalization of the data within the network in each layer has to be aligned with the covariance matrix. This matrix finds the similarities and correlation value between the training parameters values, and once there is a negative standardization value here, the normalization is required. Since training is often mostly performed with mini-batches, the covariance and mean value can be determined and used to normalize the activation in the network. Since it is possible that the mini-batch size is smaller than the number of parameters in the layer whose activation are to be normalized, a singular covariance matrix is generated. It is therefore proposed to use the variance $\sigma^2$ to assign the activation vectors parameter for normalization. The noprmalized input data $\hat{x}_i$:

\begin{equation} 
         \hat{x}_i= \frac{{x}_i- \mu}{\sqrt{\sigma^2+\in }}\
         \label{equation:Equation11}
\end{equation}

During training, the $\mu$ and the $\sigma^2$ are computed for the mini batch to determine the best values during testing. Since the normalization is carried out before the linearity or non-linearity of the activation functions, it can lead to the fact that the input variables are only in the linear part of the function. To prevent this behavior, there are two additional learnable parameters, $\gamma$ and $\beta$, which are introduced to ensure numerical stability. In the last step of the batch normalization algorithm, the computed value of $\hat{x}_i$ is shifted and scaled with the $\gamma$ \cite{Ioffe}.

\begin{equation}
	BN_{\gamma,\beta}(\vec{x}_i)=\gamma.\hat{x}_i+\beta
    \label{equation:Equation 12}
\end{equation}

The adjustment of the parameters is seamlessly integrated into the back-propagation algorithm:

\begin{equation}
	\Delta \gamma=-\gamma . \frac{\vartheta E}{\vartheta_{\gamma}}
     \label{equation:Equation 13}
\end{equation}

\begin{equation}
	\Delta \beta=-\gamma . \frac{\vartheta E}{\vartheta \beta}
     \label{equation:Equation 14}
\end{equation}

The added parameters in (\ref{equation:Equation 12}) increase the stability of the normalization process.

\section{The Proposed Adaptive BN}
\label{section4}
\begin{figure}
\includegraphics[width=1\linewidth]{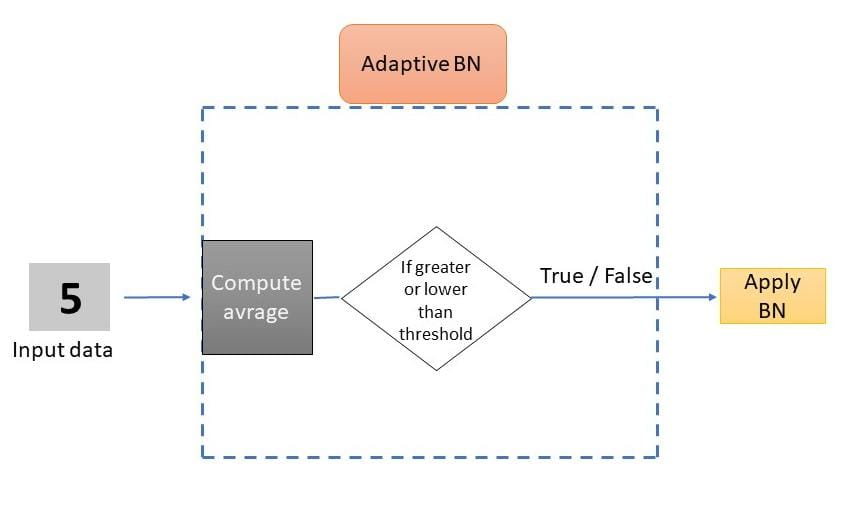}
	\caption{The Proposed Adaptive BN Method.}
	\label{fig:modelcnn}
\end{figure}

Fig. \ref{fig:modelcnn} shows the adaptive BN stage. The adaptive BN stage aims to increase the accuracy of the deep CNN data classification model by modifying the traditional BN. The adaptive BN layer turns the original BN functionality on and off. The upper and lower bound requirements are set based on a threshold which determines the need for using the BN.

In the first epoch, the pixel density is computed for each feature $V$. Let us represent each training steps as a vector of data.
\begin {equation}
V=\left\{v_1,v_2,...,v_n\right\}
\end{equation} 
where each vector $v$ is a group of features $f$.
\begin{equation}
v=\left\{f_1,f_2,f_3,....f_n\right\}
\end{equation}

The vector average is then recorded and assigned to the label category based on this information. An average computation is recorded for each category of the training data.

After recording the average for each category, the upper and lower thresholds for each category are computed by adding or subtracting the pre-assumed percentage of the value of each average of the category. These thresholds represent the range for each category.

After computing the upper and lower thresholds, the second epoch of the model starts. The mean of any feature is compared against the threshold of its class. If it is greater or less than the threshold bound, the BN is applied since the instances within the batch are out of the category range. All the training data in each subsequent epochs are passed over the computed thresholds in the first epoch.

For example, Let us suppose that we have the following average list, which represents class labels for the objects dataset:

Let $Y$ is a class label for different objects such that {$Y=\{Y_{1}, Y_{2}, Y_{3},...,Y_{n}\}$}. 
Let $\bar{A}_v$ be best average features of the input instances such that is $\bar{A}_v=\{\bar{A}_{v_1}, \bar{A}_{v_2},\bar{A}_{v_3},...,\bar{A}_{v_n}\}$. An upper percentage number $UPR-P$ is initialized. Then, we can compute the upper threshold value as follows:
\begin{equation}
{
   A_{max} = \bar{A}_v +\bar{A}v*UPR-P }
     \label{equation:Equation 15}
\end{equation}

This means that if we have a new image from the same class, which enters the training process and obtains an average pixel density, if the average pixels are more than the $upperValue$, it requires the BN to work on it. A lower percentage number $LOR-P$ is initialized. We also have a lower threshold value, which computed as follows:

 \begin{equation}
{
	A_{min} = \bar{A}_v - \bar{A}_v*LOR-P }
     \label{equation:Equation 16}
\end{equation}
If it is less than $A_{min}$, BN is needed. In this way, a swap-in and swap-out allowance are occurred rather than fixed application or exclusion of BN on the entire training dataset.


\begin{algorithm}
\caption{Adaptive BN}\label{algor}
\begin{algorithmic}[1]
\footnotesize
\renewcommand{\algorithmicrequire}{\textbf{Input:}}
\renewcommand{\algorithmicensure}{\textbf{Output:}}
\REQUIRE Training Batches, $A_{max}$, $A_{min}$ \\

\ENSURE BN; \# A Boolean decsion of batch normalization.\\ 

             \STATE\textit{{UPR-P}$\leftarrow$
             \textit{added value from the best class threshold to form the upper BN  threshold.}}\\
             \STATE\textit{{LOR-P}$\leftarrow$
             \textit{deducted value from the best class threshold to form the lower BN  threshold}}\\

             \STATE\textit{$\bar{A}_{v}\leftarrow$ \textit{Best Average Class Threshold}}\\
             \STATE\textit{\textbf{Procedure}}
        \IF {{$Epoch Count=1$}}
           \FOR {all learning steps in first epoch}
             \STATE $\bar{A}_{v}=sum(Average Features)/|Features|$
            \ENDFOR
        \ELSE
            \FOR{ $i\in[0,N_{batches},i \in \mathcal{N}] $}
        \STATE $A_{i\_max}= \bar{A}_{v\_i}+\bar{A}_{v\_i}*UPR-P$\\
        
        \STATE  $A_{i\_min}=\bar{A}_{v\_i}-\bar{A}_{v\_i}*LOR-P$\\
        
               \IF{$averageFeature> A_{i\_max} || averageFeature < A_{i\_min}$}
                 \STATE $BN_i$ = \TRUE
                \ELSE
                  \STATE$BN_i$ = \FALSE
\ENDIF
\ENDFOR
\ENDIF
\RETURN BN

\STATE \textbf{End Procedure}
\end{algorithmic}
\end{algorithm}

Algorithm~\ref{algor} shows the steps of the proposed methodology. The $UPR-P$ is the added ratio to $\bar{A}_v$  of the class to configure the upper BN threshold. This value is directly related to the accuracy of the model. The $LOR-P$ is the discounted ratio of the class's best average to set the lowest BN threshold. This value has a direct relationship to the model's accuracy. The $averageFeature$ is the average feature value of the input instance. The average feature for a data instance is the sum of the feature values divided by the numbers of the feature values. $\bar{A}_v$ is the average value extracted from the sum of the features of the data entered the training process in the first epoch from the same class. $A_{max}$ is the upper threshold value after adding the $UPR-P$ ratio to the $\bar{A}_v$. The $A_{min}$ is the lower threshold value after decreasing the $LOR-P$ ratio from $\bar{A}_v$.

In the proposed method, the adaptive BN decision stage have been added as shown in Fig. \ref{fig:modelcnn}. As shown in Table \ref{table5}, the hybrid parameters are used in the proposed adaptive BN.


\begin{table}
\caption{adaptive BN layer in a deep CNN architecture.} 
\label{table5}
\centering 
\setlength{\tabcolsep}{4pt}
\begin{tabular}{c c} 
\hline\hline 

Layer & Hybrid parameters tuning

 \\ [0.8ex] 
\hline 
Input& 32*32(width*length) \\ 
\textbf{adaptive} & \textbf{methodology layer} \\
Conv2D& kernel size 3*3\\
MaxPool & kernel size 2*2\\
Conv2D & kernel size 3*3\\
MaxPool & kernel size 2*2\\
Conv2D & kernel size 3*3\\
MaxPool & kernel size 2*2\\
Flatting  \\
Dropout & kernel size 0.2\\
Dense(softmax) & number of classes\\

\hline \hline 
\end{tabular}
\end{table}

\section{Experiments and Results Evaluations}
\label{section5} In this section, we first present the experiments, the used datasets, and evaluations. Then, we present and discuss the results.

In the experiments, we use four popular datasets to train the CNN network. The first dataset is MNIST \cite{LeCunmMNist}. The MNIST dataset includes a handwritten set of numbers for 10 classes. The second dataset is Fashion-MNIST \cite{ZalandoFashion}, which contains 70,000 images of different outfits that represent 10 categories. The third dataset is CIFAR-10 \cite{Krizhevskycifar}, which includes images from different objects representing 10 categories. The last dataset is CIFAR-100 \cite{Krizhevskycifar}, which is an expanded collection of CIFAR-10 dataset and has 100 categories.

The experiments are performed on a GPU. Google Colaboratory (Colab) and Keras 3.4.3 with TensorFlow 2 are used to implement the proposed approach. A python programming language version 3.7 is used. The code is publicly available at \url{https://github.com/waelassobhi/An-Adaptive-Batch-Normalization-In-Deep-learning}.

We perform the experiments on three different scenarios :
\begin{itemize}
\item With BN.
\item Without BN.
\item With Adaptive BN.
\end{itemize}
The four datasets we used are applied to all three
scenarios using four different batch sizes 4, 8, 16, and 32.
\newline
The metric used to compare performance between the three different models with the four bases of evidence and the four different batch sizes is the measurement of accuracy:
\begin{equation}
     Accuracy = \frac{T_P + T_N}{T_P + F_P + F_N + T_N}
     \label{equation:Equation17}
\end{equation}
To represent this measurement, a K-fold cross-validation is used, where K is 3. Then, we compute the mean accuracy and the standard deviation for each batch size. 

\subsection{MNIST DATASET}
In Table \ref{table1}, the three scenarios are also perform on the MNIST dataset using the upper and lower thresholds and different batch sizes. From Table \ref{table1}, the adaptive based approach is superior in accuracy with the exception of the superiority of without BN scenario at a batch size 32.

\begin{table}
\caption{Evaluations on MNIST dataset.} 
\label{table1}
\centering 
\setlength{\tabcolsep}{4pt}
\begin{tabular}{c c c c} 
\hline\hline 
Batch size & BN & Without BN & Adaptive BN 
 \\ [0.8ex] 
\hline 
4& 95.22\%(+/-0.85)& 95.32\%(+/- 0.93)&\textbf{96.07\%(+/-0.09)} \\

8 & 95.83\%(+/- 0.54) & 96.30\% (+/-0.70)& \textbf{96.40\% (+/-0.21)}\\ 

16 & 97.32\%(+/-0.14)& 97.37\% (+/-0.16)&\textbf{97.40\% (+/-0.24)}\\

32 & 97.70\%(+/-0.35) & \textbf{97.98\% (+/-0.10)}& 97.84\% (+/-0.11) \\

\hline \hline 
\end{tabular}
\end{table}

\begin{figure}
	\centering
        \includegraphics[width=1.05\linewidth]{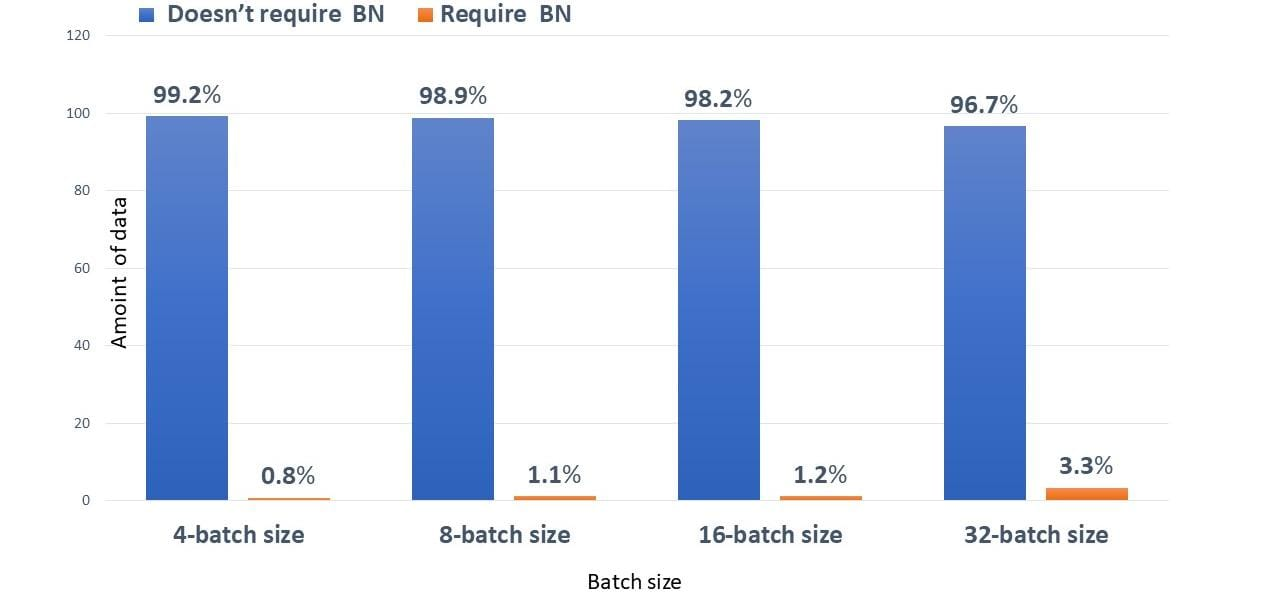}
	\caption{The amount of data don’t require the BN versus the amount of data that require the BN using the MNIST dataset.}
	\label{fig:dm2}
\end{figure}

In the experiments, the percentage of batches that are required to activate the BN as shown in Fig. \ref{fig:dm2}. We note that there is a large number of data that are passed to the training process without requiring the BN. We also note a slight increase in the percentage of batches that are required for a BN as the batch size increases.

\subsection{Fashion-MNIST DATASET}

In Table \ref{table2}, the three scenarios are also performed on the Fashion-MNIST dataset with different values of upper and lower thresholds between different batch sizes.
We can justify the different threshold values since the increase in the threshold values provides more space in the range of each category, thus introducing very few BN batches. If the threshold value is lowered, the average range is reduced. Therefore, the BN is applied to a larger number of batches. Based on these experiments, the number of batches entering the BN are controlled as these thresholds are sensitive and may obviously affect the performance of the model. It can be seen that our methodology excels in accuracy, with the exception of the superiority of the BN scenario at a batch size is 32.

\begin{table}
\caption{Evaluations on Fashion-MNIST dataset.} 
\label{table2}
\centering 
\setlength{\tabcolsep}{4pt}
\begin{tabular}{c c c c} 
\hline\hline 
Batch size & BN & Without BN & Adaptive BN 
 \\ [0.8ex] 
\hline 
4& 80.88\% (+/-1.59)& 81.11\% (+/-1.16)& \textbf{ 81.30\% (+/-0.54)}\\

8& 83.35\% (+/-0.77)& 82.11\% (+/-0.31)& \textbf{83.37\% (+/-0.78)} \\

16& 83.96\% (+/-1.33)& 84.59\% (+/-0.93)&  \textbf{84.75\% (+/-0.36)} \\

32& \textbf{86.95\% (+/-0.31)} & 85.67\% (+/-0.37)& 85.40\% (+/-0.35)\\
\hline \hline
\end{tabular}
\label{table}
\end{table}

\begin{figure}
	\centering
	\includegraphics[width=1.05\linewidth]{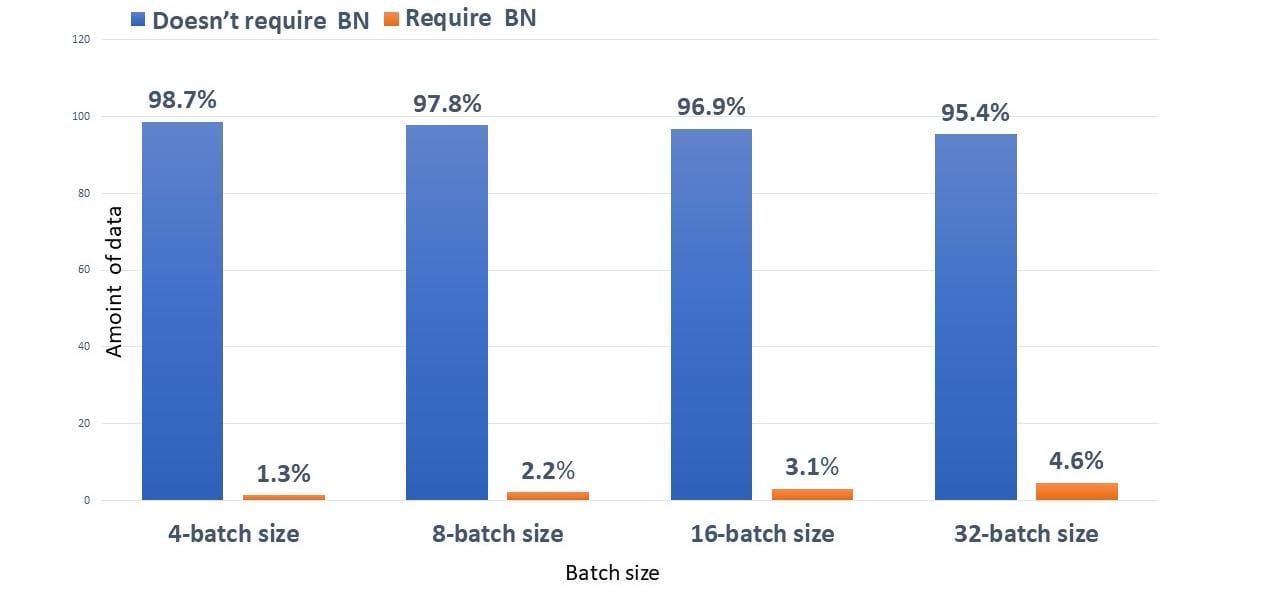}
        \caption{The amount of data don’t require the BN versus the amount of data that require the BN using the Fashion-MNIST dataset.}
	\label{fig:fm2}
\end{figure}

The percentage of batches that are required for BN is computed in Fig. \ref{fig:fm2}. We note that there is a large number of data that are passed to training without requiring the BN. We also note a slight increase in the percentage of batches require the BN activation as the batch size increases.

\subsection{CIFAR-10 DATASETS}
In Table \ref{table3}, the three scenarios are performed on the CIFAR-10 dataset with the upper and lower thresholds using different batch sizes. We can justify the different threshold values since the increase in the threshold values provides more space in the range of each category, thus introducing very few BN batches. If the threshold value is lowered, the average range is reduced. Therefore, the BN is applied to a larger number of batches. Based on these experiments, the number of batches entering the BN are controlled as these thresholds are sensitive and may obviously affect on the accuracy of the model. It can be seen that our method excels in accuracy, except for the BN scenario when the batch size is 32.

\begin{table}
\caption{Evaluations on CIFAR-10 dataset.} 
\label{table3}
\centering 
\setlength{\tabcolsep}{4pt}
\begin{tabular}{c c c c} 
\hline\hline 
Batch size & BN & Without BN & Adaptive BN 
 \\ [0.8ex] 
\hline 

4& 38.70\% (+/-1.93)& 46.16\% ( +/-1.84)& \textbf{46.72\% (+/-0.56)}\\

8& 49.42\% (+/-2.65)& 50.03\% (+/-1.83)& \textbf{51.15\% 
 (+/-0.41)}\\

16& 47.75\% (+/-5.61)& 54.74\% (+/-0.76)& \textbf{54.94\% (+/-1.06)}\\

32& \textbf{57.98\% (+/-1.68)}& 57.53\% (+/-1.59)& 57.77\% (+/-0.97) \\
\hline \hline 
\end{tabular}
\end{table}



\begin{figure}
	\centering
	\includegraphics[width=1.1\linewidth]{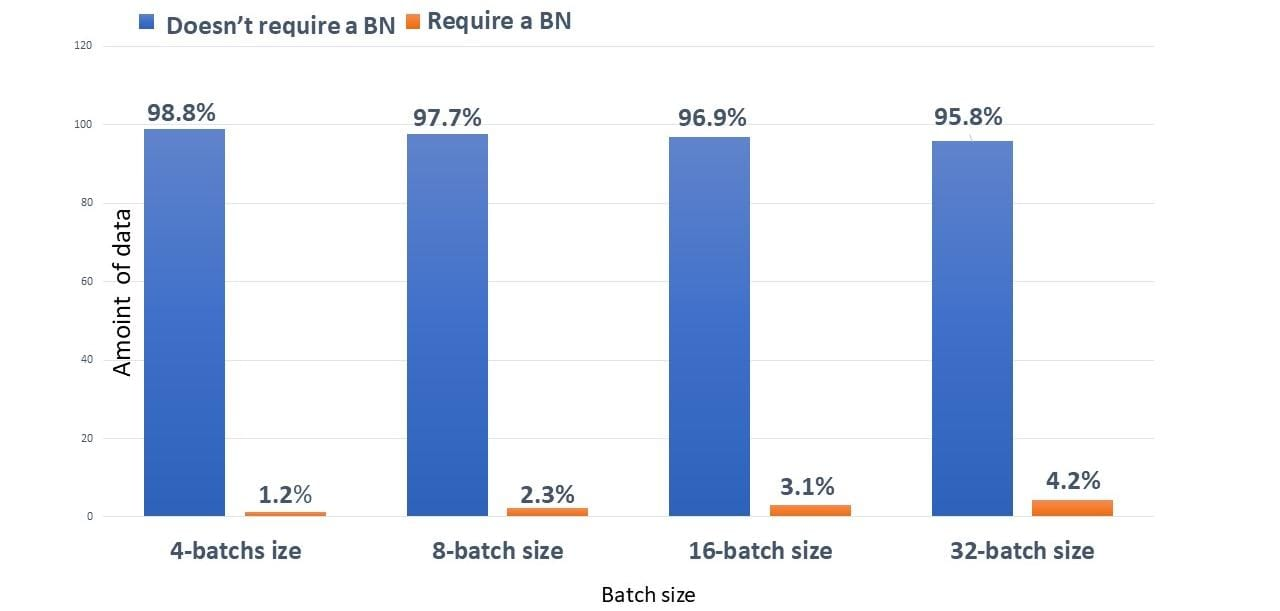}
	\caption{The amount of data don’t require a BN versus the amount of data that require the BN using the CIFAR-10 dataset.}
	\label{fig:cf102}
\end{figure}

The percentage of batches that are required for the BN is computed as shown in Fig. \ref{fig:cf102}. We note that there is a large number of data that are passed to training process without activating the BN. We also note a slight increase in the percentage of batches that require the BN as the batch size increases.

\subsection{CIFAR-100 DATASETS}
In Table \ref{table4}, the three scenarios are performed using the CIFAR-100 dataset. Upper and lower thresholds for different batch sizes are applied. It can be seen that our method is superior in accuracy using all batch sizes.

\begin{table}
\caption{Evaluations on CIFAR-100 dataset.} 
\label{table4}
\centering 
\setlength{\tabcolsep}{4pt}
\begin{tabular}{c c c c} 
\hline\hline 
Batch size & BN & Without BN & Adaptive BN 
 \\ [0.8ex] 
\hline 

4& 25.50\% (+/- 0.12)& 26.69\% (+/-1.06)& \textbf{27.12\% (+/-0.27)}\\

8& 27.18\% (+/-2.02)& 28.20\% (+/-0.94)& \textbf{28.71\% (+/-1.33)}\\

16& 27.68\% (+/-1.48)& 28.05\% (+/-0.51)& \textbf{28.28\% (+/-0.02)}\\

32& 28.89\% (+/-0.12)& 30.78\% (+/-0.94)& \textbf{31.13\% (+/-0.16)} \\

\hline \hline 
\end{tabular}
\end{table}

\begin{figure}
	\centering
        \includegraphics[width=1.1\linewidth]{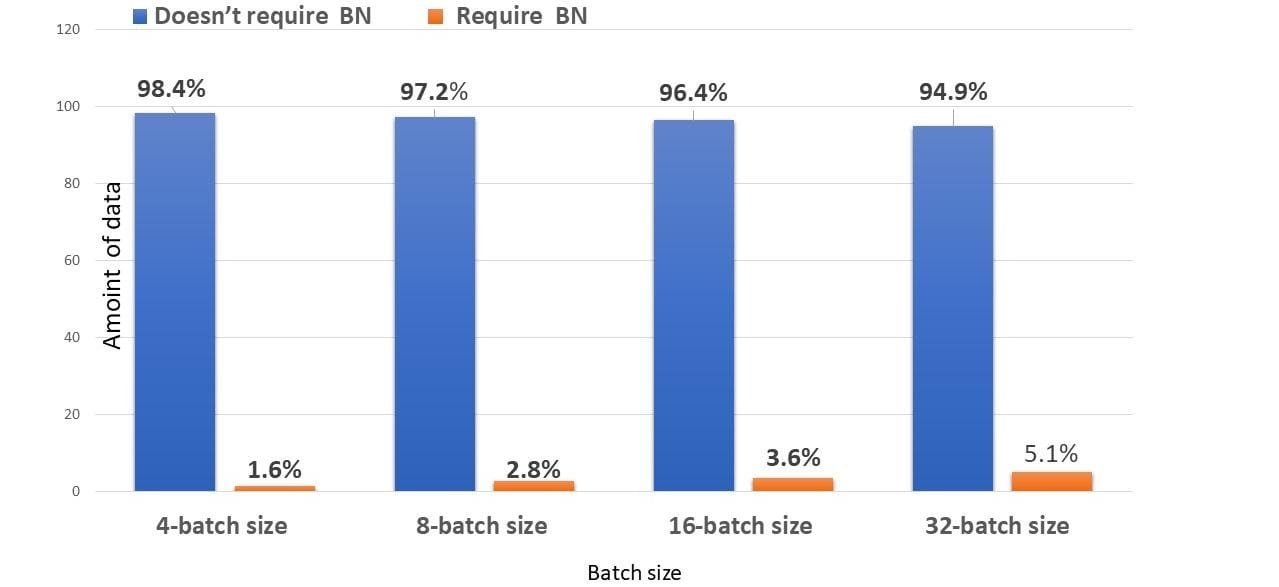}
	\caption{The amount of data don’t require the BN versus the amount of data that require the BN using the CIFAR-100 dataset.}
	\label{fig:cf1002}
\end{figure}

The percentage of batches that are required for the BN activation is computed as shown in Fig. \ref{fig:cf1002}. We note that there is a large number of data that are passed to training process without requiring the BN. We also note a slight increase in the percentage of batches that require the BN activation as the batch size increases.

\subsection{Discussion}

Working on the results to strengthen the theory that some, not all, training images require the BN. The BN is extracted since the data are not relevant to their categories.

\begin{figure}
	\centering
	\includegraphics[width=1\linewidth]{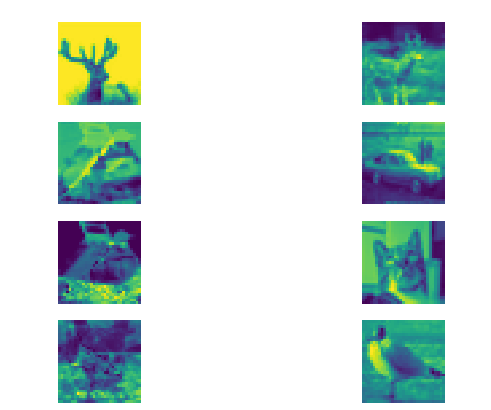}
	\caption{Sample of training images that require the BN on the left column versus a sample of the same categories on the right column that don't require the BN.}
	\label{fig:t1cifar}
\end{figure}

From Fig. \ref{fig:t1cifar}, we see different training images from the same categories. This is what we assumed, and this is what the methodology is built on, as these images prove it. Only the distorted images on the left column require to be normalized to make them close to their counterparts in the same category on the right column. We can also see the difference in pixel values between the images on both columns using the different thresholds. 
One of the main weaknesses in the methodology is determining the upper and lower bounds. The experiments find out the best threshold based on batch size, data quality, and others. Second, if there is one training image in the batch that requires to be normalized, then the entire batch are applied to the BN. Lastly, the methodology is promising 
and effective, but it results in higher accuracy mostly in small batch sizes.

\section{Conclusions and Future Work}
\label{section6}
In this paper, we proposed a threshold-based adaptive BN approach that segregates the data that needs the BN from the data that does not. The proposed approach achieves better performance than the traditional BN in most cases whose small batch sizes. It also reduces the occurrence of internal variable transformation to increase network stability.

In the future, enhancing adaptive thresholding will be based on finding a mechanism to select the best threshold values based on the behaviour of the model during the training.

\end{document}